\documentclass{article}
\usepackage{spconf,amsmath,graphicx}

\usepackage{graphicx}
\usepackage{amsmath}
\usepackage{amsthm}
\usepackage{booktabs}
\usepackage{algorithm}
\usepackage{algorithmic}
\usepackage{multirow}
\usepackage{subfigure} 

\title{Training Stronger Spiking Neural Networks with Biomimetic Adaptive Internal Association Neurons}
\name{Haibo Shen\textsuperscript{1}, 
Yihao Luo\textsuperscript{2,1}, 
Xiang Cao\textsuperscript{3,1}, 
Liangqi Zhang\textsuperscript{1}, 
Juyu Xiao\textsuperscript{1}, 
Tianjiang Wang\textsuperscript{1}\thanks{This work was supported in part by the National Natural Science Foundation of China under Grant 61572214 and Seed Foundation of Huazhong University of Science and
Technology (2020kfyXGYJ114). (Corresponding author: Tianjiang Wang.)}}
\address{School of Huazhong University of Science and Technology\textsuperscript{1}\\
Yichang Testing Technique Research Institute\textsuperscript{2}\\
Changsha University\textsuperscript{3}}
\begin{document}
%
\vspace{-10pt}
\maketitle
\begin{abstract}
As the third generation of neural networks,  
spiking neural networks (SNNs) are dedicated to 
exploring more insightful neural mechanisms
to achieve near-biological intelligence. 
Intuitively, biomimetic mechanisms are 
crucial to understanding and improving SNNs. 
For example, the associative long-term potentiation (ALTP) phenomenon 
suggests that in addition to learning mechanisms 
between neurons, there are associative effects within neurons.
However, most existing methods only focus on the former 
and lack exploration of the internal association effects.
In this paper, we propose a 
novel Adaptive Internal Association~(AIA) neuron model
to establish previously ignored influences within neurons.
Consistent with the ALTP phenomenon, 
the AIA neuron model is adaptive to input stimuli, 
and internal associative learning occurs only when both 
dendrites are stimulated at the same time. 
In addition, we employ weighted weights to measure 
internal associations and introduce intermediate caches 
to reduce the volatility of associations. 
Extensive experiments on prevailing neuromorphic datasets 
show that the proposed method 
can potentiate or depress the firing of spikes more specifically, 
resulting in better performance with fewer spikes.
It is worth noting that without adding any parameters 
at inference, the AIA model
achieves state-of-the-art performance on DVS-CIFAR10~(83.9\%) 
and N-CARS~(95.64\%) datasets.
\end{abstract} 
\vspace{-10pt}
\section{Introduction}
Inspired by the learning mechanisms of the mammalian brain, 
spiking neural networks (SNNs) are considered a promising model 
for artificial intelligence (AI) and theoretical neuroscience~\cite{2019Towards}. 
In theory, as the third generation of neural networks, 
SNNs are computationally more powerful than 
traditional artificial neural networks (ANNs)~\cite{1997Networks,ZhangZJW022}.

Essentially, SNNs are dedicated to 
exploring more insightful neural mechanisms
to achieve near-biological intelligence.
The most representative ones are the leaky integrate-and-fire~(LIF) 
model~\cite{meng2022training,guo2022recdis} and spike-timing-dependent plasticity~(STDP) 
rules~\cite{2019Towards}.
The LIF neuron model is a trade-off between biomimicry and computability. 
It can reflect most properties of biological neurons, 
while the calculation is relatively simple. 
STDP is a temporally asymmetric form of Hebbian learning that arises from 
the close temporal correlation between the spikes of two neurons, 
presynaptic and postsynaptic~\cite{ZhangZJW022}.
Both of them inspire us to understand and improve SNNs 
from biological mechanisms.
\begin{figure}[tbp]
    \centerline{\includegraphics[scale=0.5]{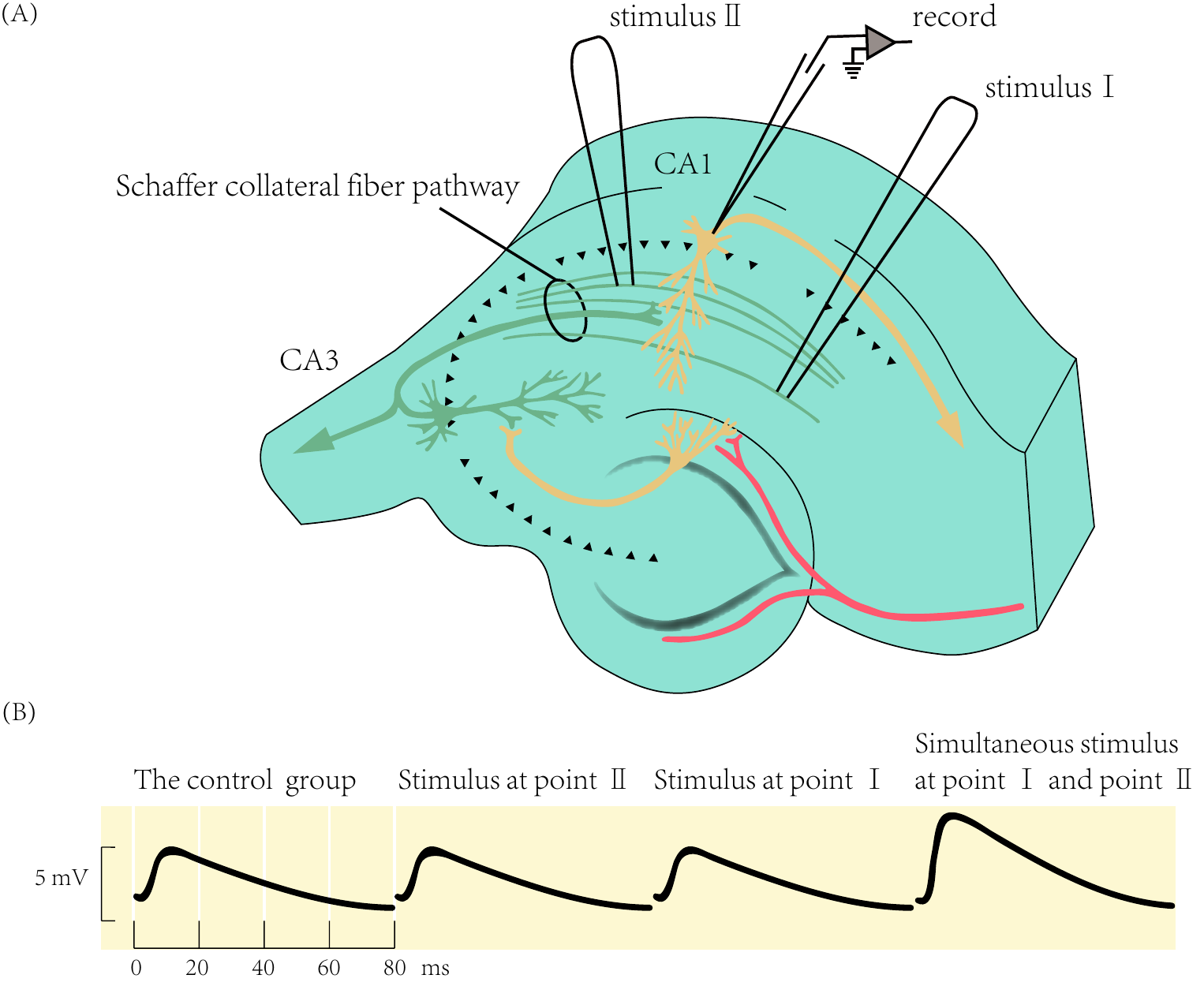}}
    \caption{The associative long-term potentiation phenomenon in hippocampal slices of rats.
    (\textbf{A}) Two different groups of Schaffer collateral presynaptic fibers 
    (stimulation $\text{I}$ and stimulation $\text{II}$) are stimulated while performing intracellular 
    recording of CA1 (cornu Ammonis) pyramidal cells.
    (\textbf{B}) The average response of CA1 pyramidal cells to a stimulus from point $\text{II}$ is observed.
    The curve depicts the membrane voltage change after point $\text{I}$ and point $\text{II}$ 
    are stimulated separately and simultaneously for ten minutes under control conditions
    \protect~\cite{Barrionuevo1983}.}\label{fig_APBS}
\end{figure} 
 
The interesting associative long-term potentiation (ALTP)
phenomenon~\cite{Barrionuevo1983} has recently 
attracted our attention, as shown in Fig.~\ref{fig_APBS}~(A).
When points $I$ and $II$ were stimulated 
at the same time, the response of point $II$ was enhanced, 
but not when stimulated separately.
The ALTP phenomenon suggests that in addition to learning mechanisms 
between neurons, there are associative effects within neurons.

However, most biologically inspired methods mainly 
focus on the learning between neurons and 
lack the exploration of associative effects within neurons.
For example, spatiotemporal backpropagation based method~\cite{zheng2021going} with surrogate gradient, 
time surfaces based method~\cite{Sironi2018} to process the temporal information of SNNs,
membrane potential based method~\cite{guo2022recdis} to rectify the distribution,
membrane time constant based method~\cite{fang2021incorporating} 
to avoid manual adjustment of neuron parameters,
axonal delay based method~\cite{SunZB22} to simulate a short term memory,
and other novel method~\cite{brainZhang2018,meng2022training}. 

In this paper, we propose an adaptive internal associative~(AIA) 
neuron model to establish neglected associations within neurons. 
Consistent with the ALTP phenomenon, 
the AIA neuron model is adaptive to input stimuli, 
and internal associative learning occurs only when both 
dendrites are stimulated simultaneously.  
Furthermore, we measure the strength of internal associations 
by weighting gradients passed between neurons 
with input weights.
Therefore, the effect is positively correlated with input weights,
as described in Eq~\ref{eq_grad_AIAWV3}.
In addition, 
we introduce intermediate caching to reduce the 
volatility of associations. 
Extensive experiments
demonstrate that the AIA neuron model can 
potentiate or depress the firing of spikes more specifically. 
This biomimetic model performs better with 
fewer spikes, which is closer to the efficiency of the brain.
It is worth noting that without adding any parameters at inference, 
the AIA model achieves state-of-the-art performance on DVS-CIFAR10~(83.9\%) 
and N-CARS~(95.64\%) datasets.

The closest works to the AIA model are the learning rules 
of Hebbian and STDP~\cite{2019Towards}. However, 
they both only focus on the learning mechanism between neurons, 
while AIA establishes the associations inside neurons.

\vspace{-5pt}
\section{Method}

\subsection{LIF Neuron Model}\label{neuron model}
For spiking neurons,
the neuronic membrane potential increases with the accumulation of weighted spikes, and
an output spike is generated once the membrane potential exceeds a threshold. 
The membrane potential of widely used LIF model~\cite{meng2022training} 
is formulated as:
\begin{equation}\label{eq_neuron_ori}
    \begin{aligned}
        \tau_{m} \frac{du}{dt}=-\left(u-V_{rest}\right)+R_{m} \cdot I(t), \quad u<V_{th} \\
        \left.
            \begin{aligned}
            u = V_{rest}&  \\
            o = \delta(t-t_i)&
            \end{aligned}
        \right\},  \quad u \geq V_{th}
    \end{aligned}
\end{equation}
where $\tau_{m}=R_{m}C_{m}$ is the membrane time constant, $R_{m}$ and $C_{m}$ are resistance constant and capacitance constant, respectively.
$u$ is the membrane potential, $I(t)$ is the input current, $V_{th}$ and $V_{rest}$ are the spiking threshold and resting potential.
Once $u$ reaches $V_{th}$ at $t_i$, a spike is generated and $u$ is reset to $V_{rest}$, which is usually taken as 0. 
The output spike $o$ is described by the Dirac delta function $\delta(x)$. The input will be summed to $I(t)$ by the dendrite response weight $w$. 
By absorbing the $1-dt/\tau_{m}$ and $1/C_{m}$ constants into response weights $w$ 
and leakage coefficient $\lambda$ respectively~\cite{zheng2021going},
the discrete form of Eq.~\ref{eq_neuron_ori} is described as:
\begin{equation}\label{eq_neuron_discrete}
    \begin{aligned}
        u_{i}^{n+1,t+1}=&\lambda u_{i}^{n+1,t} (1 - o_{i}^{n+1,t}) + x_{i}^{n+1,t+1}\\
        x_{i}^{n+1,t+1} =&\sum_{j} w_{ij}^{n+1} o_{j}^{n,t+1} \\
        o_{i}^{n+1,t+1} =&H(u_{i}^{n+1,t+1}-V_{th}) 
    \end{aligned}
\end{equation}
where n denotes the $n$-th layer and
$w_{ij}$ is the synaptic weight from the $j$-th neuron in pre-layer $n$ to the $i$-th neuron in the post-layer $n + 1$. 
$x_{i}$ denotes the weighted input, 
and $H(x)$ is the Heaviside step function,
whose gradient is given by the surrogate method~\cite{zheng2021going}.

\begin{figure*}[tbp]
    \centerline{\includegraphics[scale=0.95]{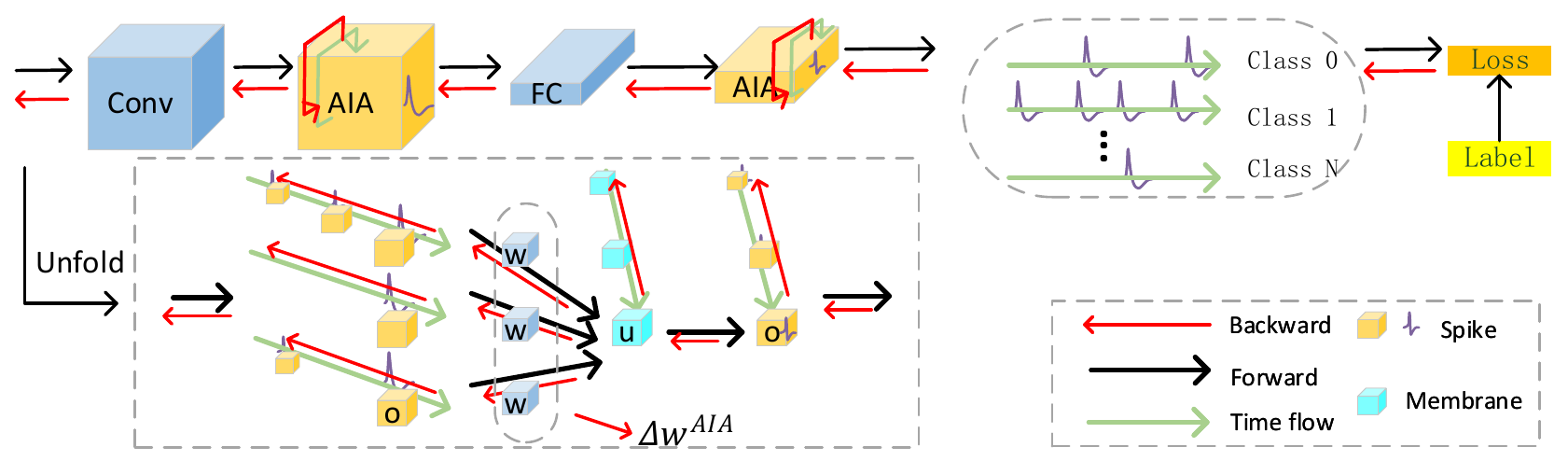}}
    \caption{The network Framework. 
    Data and gradients flow along space and time, 
    and $\Delta w^{AIA}$  is determined 
    by the input weights.}
    \label{fig_architecture}
\end{figure*}   
\vspace{-5pt}
\subsection{Adaptive Internal Association Neuron Model}\label{APBS implementation}
 
\paragraph*{Motivation} 
The phenomenon of associative long-term potentiation~\cite{Barrionuevo1983} 
in Fig.~\ref{fig_APBS} and 
the associated long-term depressive phenomenon~\cite{Stanton1989}
illustrates that
there is an associative effect within neurons when stimulated 
at the same time. 
And the effect is positively correlated with input weights.
These observations inspire us to explore the corresponding associative learning mechanisms within neurons.
\vspace{-5pt}
\paragraph*{AIA Neuron Model}
As shown in Fig.~\ref{fig_APBS}, 
between different input synapses within a single neuron, 
associative learning occurs only when both input synapses 
are stimulated. Since the biological mechanism of the ALTP 
phenomenon is unclear, we hypothesized that a function $f(x)$ 
acting on dendrites could establish stimulus-adaptive associative learning. 
We will discuss the form of this function in Sec.~\ref{sec_gradients}.
Adding the associative learning term $f(x)$ to 
Eq.~\ref{eq_neuron_discrete}, 
the adaptive internal associative~(AIA) neuron model can be formulated as:
\begin{equation}\label{eq_neuron_AIAN}
    \begin{aligned}
        u_{i}^{n+1,t+1}=&\lambda u_{i}^{n+1,t} (1 - o_{i}^{n+1,t}) + f(x_{i}^{n+1,t+1})\\
        x_{i}^{n+1,t+1} =&\sum_{j} w_{ij}^{n+1} o_{j}^{n,t+1} \\
        o_{i}^{n+1,t+1} =&H(u_{i}^{n+1,t+1}-V_{th}) 
    \end{aligned}
\end{equation} 
\subsection{Derivation of Gradients}\label{sec_gradients}
For the LIF model, the gradient of the weights is given by:
\begin{equation}\label{eq_grad_LIFW}
    \begin{aligned}
        \Delta w_{ij}^{LIF} &=
        \frac{\partial L}{\partial w_{ij}^{n+1}}
        =\frac{\partial L}{\partial u_{i}^{n+1,t+1}} 
        \frac{\partial u_{i}^{n+1,t+1}}{\partial x_{i}^{n+1,t+1}} 
        \frac{\partial x_{i}^{n+1,t+1}}{\partial w_{ij}^{n+1}}\\
        &=\frac{\partial L}{\partial u_{i}^{n+1,t+1}} 
        o_{j}^{n,t+1}
    \end{aligned}
\end{equation}
where L is the loss function.
For the AIA neuron model, the gradient of the weights is given by:
\begin{equation}\label{eq_grad_AIAWV1}
    \begin{aligned}
        \Delta w_{ij}^{AIA} 
        =\frac{\partial L}{\partial w_{ij}^{n+1}}
        &=\frac{\partial f(x_{i}^{n+1,t+1})}{\partial x_{i}^{n+1,t+1}} 
        \frac{\partial L}{\partial u_{i}^{n+1,t+1}} o_{j}^{n,t+1}
    \end{aligned}
\end{equation}
Combining the observations in motivation,
we can set the function $f'(x)=x$ to construct a 
stimulus-adaptive internal associative neuron model.
The $\Delta w_{ij}^{AIA}$ is formulated as:
\begin{equation}\label{eq_grad_AIAWV2}
    \begin{aligned}
        \Delta w_{ij}^{AIA} 
        &=\left(\sum_{k} w_{ik}^{n+1} o_{k}^{n,t+1}\right)
        \frac{\partial L}{\partial u_{i}^{n+1,t+1}} o_{j}^{n,t+1}
    \end{aligned}
\end{equation}
where $k$ represents the neuron connected to neuron $i$. Note that $o\in\{0, 1\}$, Eq.~\ref{eq_grad_AIAWV2} 
can be expressed as:   
\begin{equation}\label{eq_grad_AIAWV3}
    \begin{aligned}
        \Delta w_{ij}^{AIA} 
        &=o_{j}^{n,t+1}*\sum_{k}o_{k}^{n,t+1} 
        w_{ik}^{n+1} \Delta w_{ik}^{LIF}
    \end{aligned}
\end{equation}
When the neuron $j$ is in the resting state($o_j=0$), 
the weight $w_{ij}$ remains unchanged.
When neuron $j$ inputs a stimulus ($o_j=1$), 
all neurons that simultaneously input a stimulus ($o_k=1$) 
affect $w_{ij}$.
In other words,
only when neuron $i$ is stimulated by both 
the $j$-th and $k$-th neurons 
will the corresponding associative effects 
be established within the neuron.

In addition, Eq.~\ref{eq_grad_AIAWV3} 
measures the association influence
with minimum cost by multiplexing 
the weights and the gradients between neurons. 
It turns out that the greater the synaptic response 
to the stimulus, the stronger the effect of 
associative learning, 
which is also consistent with the observations.
Fig.~\ref{fig_architecture} shows the network framework, 
and Eq.~\ref{eq_neuron_AIAN} and Eq.~\ref{eq_grad_AIAWV3} 
represent computations in the forward and backward respectively.

\subsection{Intermediate Cache}
In practice, the gradient swings 
between $0$ and $w_{ik}^{n+1} \Delta w_{ik}^{LIF}$, 
which may cause the problem of ``dead neurons", 
so we introduce an intermediate cache variable $\beta$. 
Specifically, let $f'(x)=\beta$, the gradients of 
$\Delta w_{ij}^{AIA}$ and $\beta$ are:
\begin{equation}\label{eq_grad_AIAWV4}
        \Delta w_{ij}^{AIA}=\frac{\partial L}{\partial w_{ij}^{n+1}}
        =o_{j}^{n,t+1} *\beta \Delta w_{ij}^{LIF}
\end{equation}
\begin{equation}\label{eq_grad_AIAB}
        \frac{\partial L}{\partial \beta} = 
        \sum_{k} o_{k}^{n,t+1} w_{ik}^{n+1} \Delta w_{ik}^{LIF}
\end{equation}
It can be seen that, after adding the intermediate cache, 
$\beta$ accumulates gradients in Eq.~\ref{eq_grad_AIAWV3} 
overall stimulated synapses and applies this association effect 
only to stimulated synapses.
It is worth noting that the intermediate cache $\beta$, 
as an auxiliary tool for saving gradients, 
can be merged into $w$ for calculation at inference time, 
so it will not bring additional parameters.
 



 
\begin{table*}
    \centering 
    \caption{Performance of AIA neuron model and the SOTAs on CIFAR10-DVS, N-Caltech101 and N-CARS datasets. 
    }\label{tab_all_acc}
    \begin{tabular}{ccccccc}
    \toprule
    Network &Method& Reference & Model &CIFAR10-DVS & N-Caltech101 & N-CARS \\ \midrule
    \multirow{3}{*}{CNNs-based} & RG-CNNs~\cite{BiCABA20}    & TIP 2020 & Graph-CNN &54.00 &61.70 &91.40\\
    &           SlideGCN~\cite{LiZYZCB021}    & ICCV 2021 & Graph-CNN &68.00 &76.10 &93.10\\
    &           ECSNet~\cite{ECSNet}    & T-CSVT 2022 & LEE$\rightarrow$MA &72.70  &69.30& 94.60\\
    \midrule 
    \multirow{10}{*}{SNNs-based} & HATS\cite{Sironi2018}& CVPR 2018 & HATS-SVM & 52.40 & 64.20 & 81.0 \\
    &Dart\cite{RameshYOTZX2020} & TPAMI 2020 & SPM-SVM & 65.80 & 66.80 & -  \\ 
    &STBP~\cite{zheng2021going} & AAAI 2021 & Resnet-19 &67.80 & -& -\\ 
    &PLIF~\cite{fang2021incorporating}    & ICCV 2021 & 7-layer &74.80 & -& -\\
    &Dspike~\cite{li2021differentiable}    & NeurIPS 2021 & ResNet-18 &75.40 & -& -\\ 
    &AutoSNN~\cite{na2022autosnn}    & ICML 2022& - &72.50 & -& -\\
    &RecDis~\cite{guo2022recdis}    & CVPR 2022& Resnet-19 &72.42& -& -\\
    &DSR~\cite{meng2022training}    & CVPR 2022& VGG-11 &77.27& -& -\\ 
    &NDA~\cite{li2022neuromorphic}    & ECCV 2022& VGG-11 & 81.70& 78.20 &90.1 \\ \cmidrule(l){2-7}
    &AIA                             & - &VGG-9 & \textbf{83.90}& \textbf{80.07} &\textbf{95.64}\\ 
    \bottomrule
    \end{tabular}
\end{table*}
\section{Experiments} \label{sec_experiments}
Extensive experiments are conducted to demonstrate 
the superiority of the AIA neuron model. 
For a fair comparison, a LIF neuron model with the same parameters 
is used as the baseline.
Specifically, the weights are initialized by kaiming distribution. 
The threshold and leakage coefficients $\lambda$ of neurons 
are 1 and 0.5 respectively.
Adam optimizer is introduced to adjust the learning rate, 
which is initially set to $1 \times 10^{-3}$.

\subsection{Comparison with the State-of-the-Art}
As shown in Tab.~\ref{tab_all_acc}, we surpass previous 
state-of-the-art methods on DVS-CIFAR10~\cite{Li2017}, 
N-Caltech101~\cite{N101}, N-CARS~\cite{Sironi2018} neuromorphic datasets.
\textbf{DVS-CIFAR10} contains 10,000 samples, carrying noise and blur caused by event cameras. 
We resize the samples to $64\times64$ and achieve \textbf{83.9\%} accuracy. 
It is worth noting that NDA is the best performer in the previous work, 
but its sample size is $128\times128$.
To the best of our knowledge, the AIA neuron model achieves 
state-of-the-art performance on DVS-CIFAR10 dataset among all SNNs.
\textbf{N-Caltech101} is a spiking version of the original
frame-based Caltech101 dataset. \textbf{N-CARS} is a large 
real-world event-based car classification dataset extracted 
from various driving courses.
On these two datasets, we resize the samples to $48\times48$, 
reaching the accuracy of \textbf{80.07\%} and \textbf{95.64\%} respectively.
In conclusion, the AIA model achieves leading performance 
by establishing associations within neurons, 
which in turn illustrates the effectiveness of 
biomimetic internal associations.

\subsection{Analysis of AIA Neuron Model}
\begin{table}[b]
    \centering
    \caption{The performance of different neuron models on N-Caltech101 datasets and CIFAR10-DVS datasets.}
    \label{tab_ablation_study}
    \begin{tabular}{ccc}
    \toprule
    Neurons   & N-Caltech101(\%) & CIFAR10-DVS(\%) \\
    \midrule
    LIF         & 78.13                 & 82.5        \\
    IF          & 75.94                 & 81.9        \\
    PLIF        & 75.58                 & 79.0          \\
    AIA         & \textbf{79.59}       & \textbf{83.5}      \\
    Cached AIA  & \textbf{80.07}       & \textbf{83.9}       \\
\bottomrule
\end{tabular}
\end{table}

\paragraph*{Compared with other neuron models}
Ablation experiments are performed 
on CIFAR10-DVS and N-Caltech101 datasets to 
further compare the effects of AIA and other neuronal models. 
We employ a publicly available implementation of 
PLIF neurons~\cite{fang2021incorporating}, 
which sets the membrane time constant as a trainable variable. 
As shown in Tab.~\ref{tab_ablation_study}, 
the AIA neuron model achieves the accuracy 
of \textbf{79.59\%} and \textbf{83.5\%},
surpassing the most commonly used neuron models under the same parameters. 
In addition, the intermediate cache also improves the AIA model~(Cached AIA).

\begin{figure}[tb]
    \begin{minipage}[b]{.48\linewidth}
      \centering
      \centerline{\includegraphics[width=4.0cm]{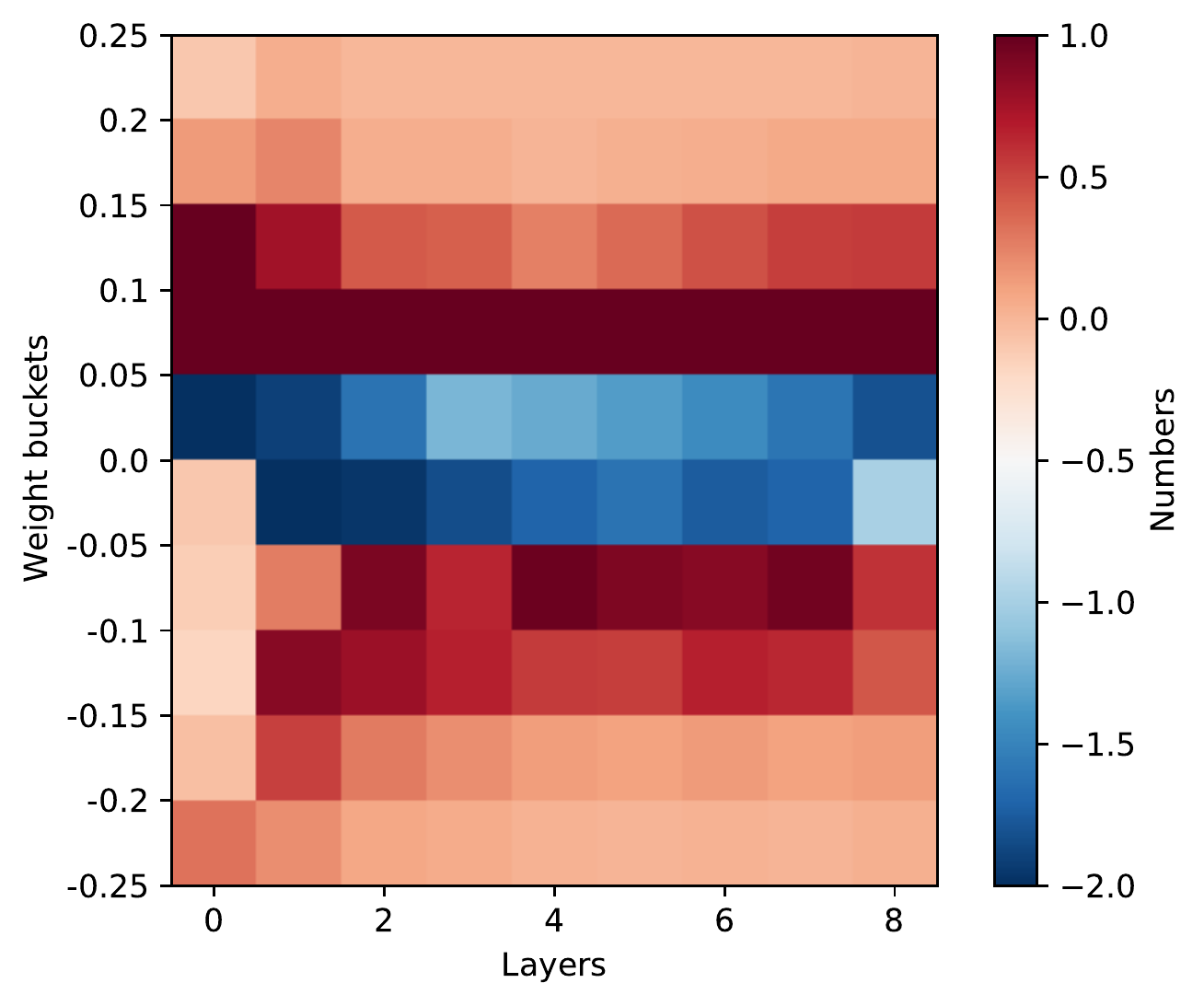}}
      \centerline{(a) Weight distribution.}\medskip
    \end{minipage}
    \hfill
    \begin{minipage}[b]{0.48\linewidth}
      \centering
      \centerline{\includegraphics[width=4.0cm]{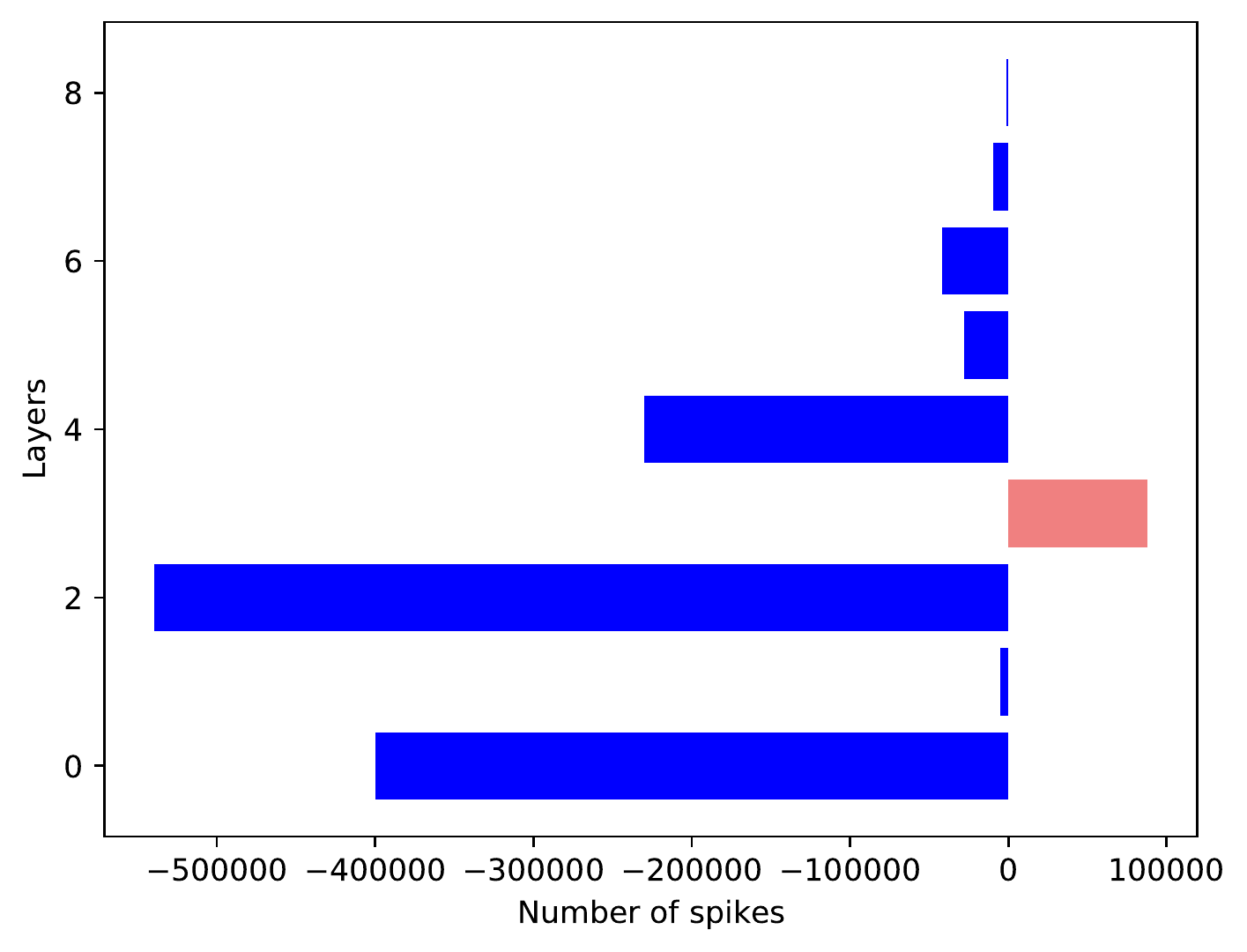}}
      \centerline{(b) Number of spikes.}\medskip
    \end{minipage}
    \caption{Variations in  
            weight distribution and number of spikes 
            after using AIA neurons 
            compared to LIF neurons.}
    \label{fig_analysis}
\end{figure}

\paragraph*{Insightful variations of the AIA model}
To gain further insight into the workings of AIA neurons, 
we perform some analysis on the N-Caltech101 dataset.
As shown in Fig.~\ref{fig_analysis} (a), 
we calculate the weight distribution of the AIA model 
and the LIF model respectively, 
and obtain the normalized changes in different intervals.
Red indicates that the number of weights 
in the interval is increasing, and blue indicates 
that it is decreasing. 
It turns out that the weights escape from intervals close to 0, 
implying a larger change in membrane voltage.
In other words, AIA neurons can more effectively potentiate or depress 
the firing of spikes, resulting in better extraction of key features. 

In addition, the change in the number of spikes is shown 
in Fig.~\ref{fig_analysis} (b). 
The AIA model significantly affects the number of fired spikes, 
especially in feature extraction layers. 
Notably, the number of fired spikes is significantly
reduced overall, implying that the AIA neuron model 
performs better with fewer spikes.
It indicates that the biomimetic internal association mechanism 
inherently improves the efficiency and biological plausibility of the model.

\section{Conclusion}
Inspired by the ALTP phenomenon, 
we propose a novel adaptive internal association~(AIA) neuron model. 
This biomimetic model achieves state-of-the-art performance
on neuromorphic datasets. In addition, 
the intermediate cache is introduced to reduce volatility. 
The insightful analysis shows that AIA neurons can more 
effectively potentiate or depress the firing of spikes
and perform better with fewer spikes.
And the proposed neuron model can provide the basis for more efficient and 
biologically plausible spiking neural networks.
\bibliographystyle{IEEEbib}
\bibliography{new}

\end{document}